\documentclass{article}

\PassOptionsToPackage{numbers, sort}{natbib}
\usepackage[dblblindworkshop,final]{neurips_2026}

\workshoptitle{Managing Agents that Manage Agents: Workshop on Responsible Use of Meta-Agents}

\usepackage[utf8]{inputenc}
\usepackage[T1]{fontenc}
\usepackage{amsmath}  
\usepackage{amssymb}
\usepackage{array}
\usepackage{booktabs}
\usepackage{float}
\usepackage{graphicx}
\usepackage{subcaption}
\usepackage{tikz}
\usepackage{wrapfig}
\usepackage{microtype}
\usepackage{xcolor}
\usepackage{url}
\usepackage{hyperref}
\usepackage{listings}

\title{Credit Where Credit Is Due: Contrastive Attribution Topology Search for Multi-Agent Systems}

\author{%
  Fabian Tritthart \\
  University of Freiburg \\
  \texttt{fabian.tritthart@email.uni-freiburg.de} \\
  \And
  Tom Zehle \\
  University of Freiburg \\
  \texttt{tom.zehle@tue.ellis.eu}
}
\begin{document}

\maketitle

\begin{abstract}
An LLM multi-agent system (MAS) is only as good as the agents it chooses to trust: its accuracy and cost hinge on its topology, meaning which agents are included and how they communicate. A single task score, however, does not reveal which agents help or hinder performance. We propose Contrastive Attribution Topology Search (CATS), which searches for compact teams within a fixed supergraph of specialized agents. An LLM attributer compares execution traces across candidate topologies and assigns credit to individual agents. CATS uses these credits to update supergraph edges while progressively pruning weak edges and agents, yielding a small team for deployment. With small open-weight models and matched candidate evaluations, CATS improves mean accuracy on BigCodeBench by 5.3 percentage points over score-only heuristic search while using 54\% fewer deployment tokens. On MBPP, CATS matches the strongest baselines with fewer tokens than the other searches. In a HotpotQA prompt-poisoning stress test, CATS excludes all adversarially prompted agents in all three runs, whereas random search retains some at comparable accuracy. Per-agent credit thus supports both team optimization and the pruning of potentially harmful agents.
\end{abstract}

\section{Introduction}
\label{sec:introduction}

LLM-based multi-agent systems split work across specialized agents, for
example one that writes code and another that tests it. Such collaboration
can improve small models~\citep{shenSmallLLMsAre2024,sunkaraneniAgenticSystemsBoosting2026},
but more agents and messages mean more tokens. Building a good system
therefore requires choosing which agents to include and how they
communicate, that is, the \emph{topology}. Since the best topology depends
on the task, manual design is
difficult~\citep{liSurveyLLMBasedMultiAgent2024,heLLMBasedMultiAgentSystems2025},
which motivates automated search over candidate teams.

Task scores, however, evaluate the team as a whole: a correct solution may
contain unnecessary work, and a failed one may include useful
contributions. Scores alone thus say little about which agents to keep.
Can feedback on individual agents guide the search for compact, effective
teams?

We introduce \emph{Contrastive Attribution Topology Search (CATS)}, an
offline meta-agent that assembles teams from a fixed agent pool. Starting
from a \emph{supergraph} of candidate links, CATS evaluates sampled teams
on target-domain tasks. An LLM compares their transcripts and assigns each
agent signed credit. CATS then updates link probabilities, prunes weak
links and unreachable agents, and selects one fixed team for deployment.
Agent models, prompts, roles, and tools never change.

Our contributions are: \emph{(i)} contrastive per-agent credit to guide
team construction (Figure~\ref{fig:cats-overview}),
\emph{(ii)} higher mean accuracy and lower deployment token use than
score-only search on two coding benchmarks with matched candidate evaluations
(Table~\ref{tab:coding-results}), and \emph{(iii)} a HotpotQA stress
test where CATS excludes all poisoned agents in every run, while random
search retains some at comparable accuracy
(Table~\ref{tab:hotpotqa-results}).

\section{Method}
\label{sec:method}

\begin{figure*}[t]
  \centering
  \includegraphics[width=\textwidth]{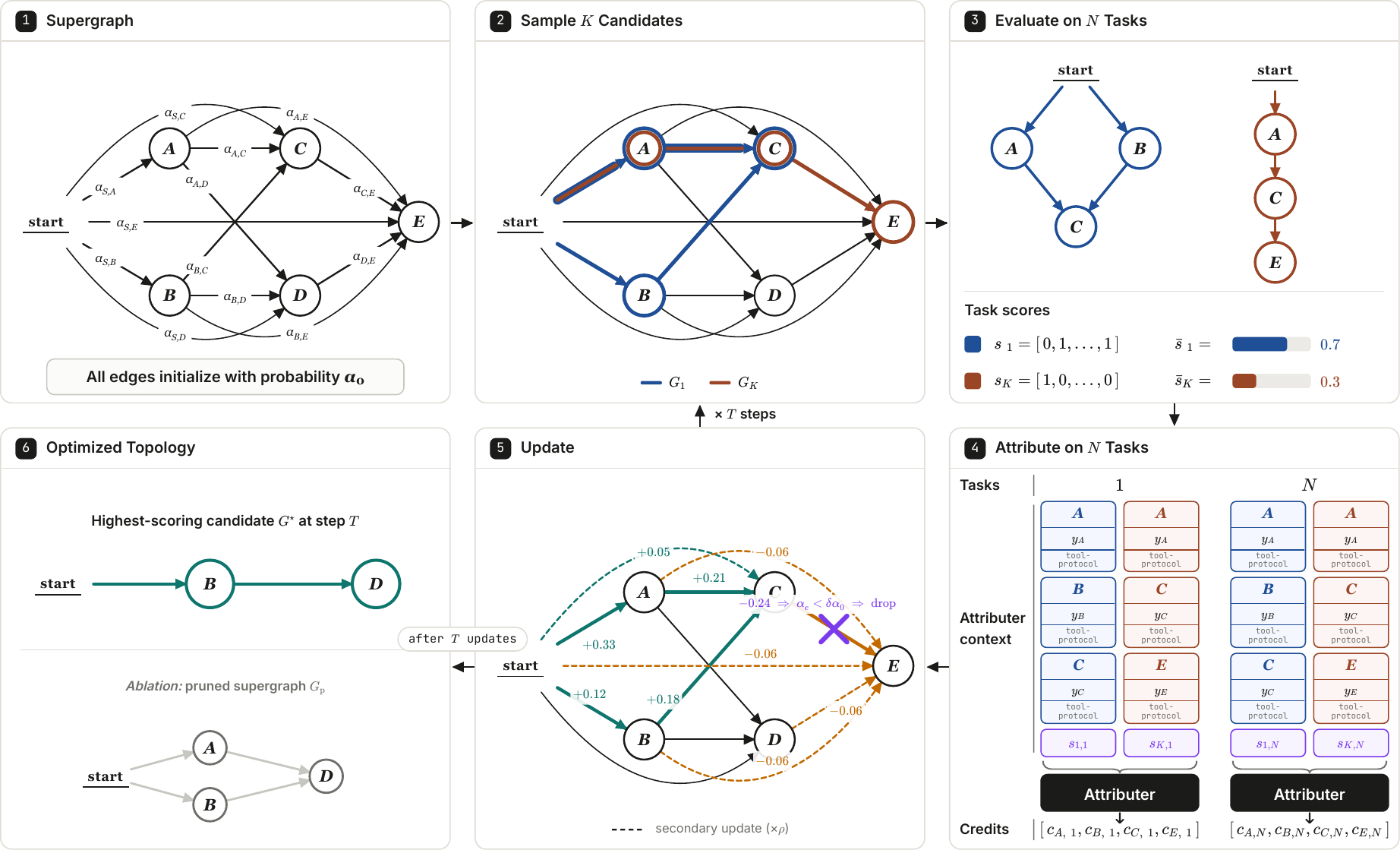}
  \caption{\textbf{CATS pipeline.} At each of $T$ steps, $K$ subgraphs
  are sampled from the supergraph and evaluated on $N$ tasks.
  The depicted supergraph is a simplified example.}
  \label{fig:cats-overview}
\end{figure*}

CATS repeatedly samples teams from a supergraph, credits each agent, and prunes weak edges.

\textbf{Supergraph.} We search for an optimized multi-agent topology as a
subgraph of a hand-crafted complete directed acyclic graph specialized to a
chosen problem domain (Figure~\ref{fig:cats-overview}). Its
nodes are agents and its edges denote direct communication (and thereby the
scheduling order).
Agents are specialized via system prompts, optional reasoning, and tool sets.
We hand-craft them using domain knowledge (automating this is left to future
work). Each edge $e$ carries a logit $\theta_e$ with sampling probability
$\alpha_e=\sigma(\theta_e)$, initialized to a common $\alpha_0$. Following
DARTS~\citep{liuDARTSDifferentiableArchitecture2019}, we relax the discrete edge choices to these
continuous parameters but replace gradient descent with heuristic credit-based
updates derived from contrastive attribution.

\textbf{Sampling.} At each step, we sample $K$ topologies by independently
drawing $e\sim\operatorname{Bernoulli}(\alpha_e)$, rejecting graphs with nodes
unreachable from the source. Since agents share an environment, not every agent needs a downstream consumer: edges are optional communication channels. We
evaluate each candidate on $N$ tasks, obtaining scores $s_{k,q}$. Candidates,
tasks, and often agents run in parallel.

\textbf{Attribution.}
An LLM attributer assigns each participating agent $i$ in candidate $k$ on task
$q$ a credit $c_{k,q,i}\in[-1,1]$ (harmful to helpful) by comparing
candidate rollouts on the same task. The attributer sees task scores,
length-capped agent outputs, tool calls, and the final artifact state
(e.g., the coding environment), and runs in parallel across tasks.

\textbf{Update.}
Credits are normalized into per-agent advantages
$A_{k,i}=|Q_{k,i}|^{-1}\sum_{q\in Q_{k,i}}(c_{k,q,i}-b_q)$, where $Q_{k,i}$ holds the tasks with a credit for $i$ in $k$ and $b_q$ is
the mean credit over participating agents and candidates for task $q$. Each edge
$(u,v)$ fired in candidate $k$ is updated by
$\theta_{uv}\leftarrow\theta_{uv}+\eta A_{k,v}$. Unfired non-source edges into an active
agent $v$ receive a smaller secondary update $\theta_{uv}\leftarrow\theta_{uv}+\eta\rho
A_{k,v}$. After each step we prune weak edges whose probability has fallen
below the fraction $\delta$ of its initial value (the \emph{pruning fraction}),
$E\leftarrow\{e\in E:\alpha_e\geq\delta\alpha_0\}$, and remove nodes that are
no longer reachable from the source. This shrinks $|E|$ and thus $|V|$, reducing the per-step cost.

\textbf{Selection.}
After $T$ steps we return the \emph{best final candidate}
$G^\star=\arg\max_{k}\tfrac{1}{N}\sum_q s_{k,q}$ over the step-$T$ candidates,
which are sampled under the accumulated priors. Ties are broken in favor of
fewer agents and then the earlier candidate. As an ablation,
we also evaluate the \emph{pruned supergraph} $G_{\mathrm{p}}$, consisting of
all nodes and edges that remain after pruning.

\textbf{Baselines.}
We compare against (i) a direct LLM call without tools, (ii) a
ReAct~\citep{yaoReActSynergizingReasoning2023} single agent, both with
literature-derived system prompts (Appendix
Table~\ref{tab:baseline-definitions}), (iii) a hand-crafted
multi-agent system (MAS)
designed before observing supergraph results (Appendix
Figure~\ref{fig:baseline-planner-executor-verifier}), (iv) random topology
search, and (v) a
score-only heuristic search using CATS's fired/unfired update and hyperparameters,
with scalar task scores replacing agent credits. Both search baselines match
$K\times T$ candidate evaluations on the same tasks, not compute
(search costs: Appendix Table~\ref{tab:search-cost}).

\section{Benchmarking}
\label{sec:evaluation}
\textbf{Experimental setup.} We target the local-deployment setting and therefore use small open-weight models: Gemma 4 12B for the MAS agents and the larger Gemma 4 31B endpoint for the attributer~\citep{gemmateamGemma4Technical2026}. We evaluate CATS on two agentic coding benchmarks, BigCodeBench and MBPP~\citep{zhuoBigCodeBenchBenchmarkingCode2025,austinProgramSynthesisLarge2021}. Each search runs for 4 optimization steps with a high learning rate and pruning fraction (Table~\ref{tab:search-hyperparameters}). This configuration performed well in preliminary experiments that excluded the held-out tasks, though we do not claim it is optimal. We repeat all experiments over three random seeds and report means with standard deviations.

\providecommand{\sd}[1]{{\scriptsize$\pm$#1}}   
\begin{figure}[t]
  \begin{minipage}[t]{0.555\textwidth}
    \centering
    \setlength{\abovecaptionskip}{0pt}
    \setlength{\belowcaptionskip}{7pt}
    \captionsetup{justification=raggedright,singlelinecheck=false}
    \captionof{table}{\textbf{Searched topologies versus hand-crafted and
    single-agent baselines.} Held-out coding tasks, mean $\pm$ sample SD over
    3 seeds, 100 tasks per benchmark. Best score per benchmark in bold.}
    \phantomsection\label{tab:coding-results}
    \footnotesize
    \setlength{\tabcolsep}{2pt}
    \begin{tabular*}{\linewidth}{@{\extracolsep{\fill}}l r@{\,}l r@{\,}l@{}}
      \toprule
      System & \multicolumn{2}{c}{Score \%\,$\uparrow$} & \multicolumn{2}{c}{Tokens/task (k)\,$\downarrow$} \\
      \midrule
      \multicolumn{5}{@{}l}{\emph{BigCodeBench}} \\
      \hspace{0.4em}\textbf{CATS} ($G^\star$) & \textbf{48.0} & \sd{5.0}  & 6.7  & \sd{1.8} \\
      \addlinespace[2pt]
      \hspace{0.4em}Direct LLM                & 29.7 & \sd{8.7}  & 0.7   & \sd{0.0}   \\
      \hspace{0.4em}ReAct single agent        & 39.7 & \sd{6.4}  & 4.7  & \sd{2.0} \\
      \hspace{0.4em}Hand-crafted MAS          & 40.3 & \sd{11.2} & 9.6  & \sd{4.5} \\
      \hspace{0.4em}Random topology search    & 35.0 & \sd{7.0}  & 8.7  & \sd{2.6} \\
      \hspace{0.4em}Score-only search& 42.7 & \sd{4.9}  & 14.7 & \sd{8.6} \\
      \addlinespace[2pt]
      \hspace{0.4em}\emph{Ablation:} pruned supergraph $G_{\mathrm{p}}$ & 45.7 & \sd{4.0}  & 13.9 & \sd{4.1} \\
      \addlinespace[3pt]
      \multicolumn{5}{@{}l}{\emph{MBPP}} \\
      \hspace{0.4em}\textbf{CATS} ($G^\star$) & \textbf{80.7} & \sd{7.5}  & 5.3  & \sd{0.7}   \\
      \addlinespace[2pt]
      \hspace{0.4em}Direct LLM                & 76.0 & \sd{6.9}  & 0.4   & \sd{0.0}    \\
      \hspace{0.4em}ReAct single agent        & 76.7 & \sd{5.0}  & 3.0  & \sd{1.1}  \\
      \hspace{0.4em}Hand-crafted MAS          & 80.3 & \sd{4.7}  & 4.2  & \sd{0.3}   \\
      \hspace{0.4em}Random topology search    & 80.3 & \sd{4.7}  & 18.8 & \sd{11.3} \\
      \hspace{0.4em}Score-only search& 79.0 & \sd{5.6}  & 22.2 & \sd{10.7} \\
      \addlinespace[2pt]
      \hspace{0.4em}\emph{Ablation:} pruned supergraph $G_{\mathrm{p}}$ & \textbf{80.7} & \sd{5.1} & 10.0 & \sd{5.3} \\
      \bottomrule
    \end{tabular*}
  
  \end{minipage}\hfill
  \begin{minipage}[t]{0.42\textwidth}
    \centering
    \captionsetup{justification=raggedright,singlelinecheck=false}
    \setlength{\abovecaptionskip}{0pt}
    \setlength{\belowcaptionskip}{7pt}
    \captionof{table}{\textbf{Poisoned-pool stress test on HotpotQA.}
    Results for CATS, random search, and score-only search (mean $\pm$ sample SD, 3 seeds, 100 tasks).}
    \phantomsection\label{tab:hotpotqa-results}
    \footnotesize
    \setlength{\tabcolsep}{2pt}
    \begin{tabular*}{\linewidth}{@{\extracolsep{\fill}}l r@{\,}l r@{\,}l r@{\,}l@{}}
      \toprule
        & \multicolumn{2}{c}{CATS} & \multicolumn{2}{c}{Random} & \multicolumn{2}{c}{Score-only} \\
      \cmidrule(lr){2-3} \cmidrule(lr){4-5} \cmidrule(l){6-7}
      Poisoned agents\,$\downarrow$ & \textbf{0.0}   & \sd{0.0}   & 2.0   & \sd{1.0}   & 1.3  & \sd{2.3} \\
      Score \%\,$\uparrow$     & \textbf{59.7}  & \sd{3.8}   & 57.7  & \sd{3.2}   & 24.3 & \sd{2.5} \\
      Tokens/task (k)\,$\downarrow$ & 12.7  & \sd{5.3}   & 15.1  & \sd{4.3}   & \textbf{3.8} & \sd{2.5} \\
      \bottomrule
    \end{tabular*}

    \vspace{10pt}

    \setlength{\abovecaptionskip}{7pt}
    \setlength{\belowcaptionskip}{0pt}
    \includegraphics[width=\linewidth]{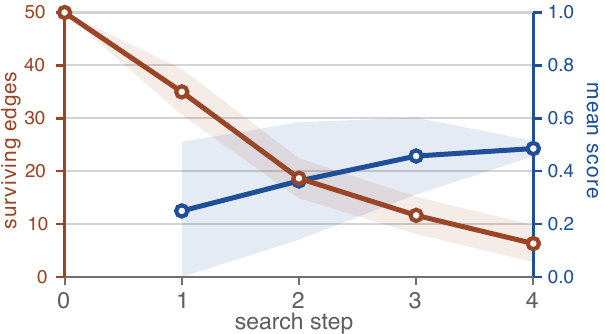}
    \captionof{figure}{\textbf{BigCodeBench search dynamics} (mean $\pm$ SD, 3 seeds).}
    \phantomsection\label{fig:bigcodebench-search-dynamics}
  
  \end{minipage}
\end{figure}

\textbf{BigCodeBench.} Our selected topology $G^\star$ improves mean accuracy
over the strongest baseline, score-only heuristic search, by 5.3 points while using
54\% fewer deployment tokens with matched candidate evaluations
(Table~\ref{tab:coding-results}). Gains hold across all three seeds,
but standard deviations overlap and uncertainty remains
(Table~\ref{tab:per-seed-results}).

\textbf{MBPP.} $G^\star$ performs comparably to the hand-crafted MAS and
random search, using fewer deployment tokens than the other searches but
more than the hand-crafted MAS (Table~\ref{tab:coding-results}).

\textbf{Final-candidate ablation.} Deploying $G^\star$ instead of
$G_{\mathrm{p}}$ roughly halves coding deployment tokens, with
higher mean accuracy on BigCodeBench and the same on MBPP
(Table~\ref{tab:coding-results}).

\textbf{Offline search tokens.} Including attribution, CATS uses 28\% and
35\% fewer mean search tokens than score-only search on BigCodeBench and
MBPP, respectively, but roughly twice those of either search baseline on
poisoned HotpotQA (Table~\ref{tab:search-cost}).

\textbf{Discovered architectures.} Every selected coding team retains a draft
coder; 3 of 6 have no inter-agent edges, but agents can still interact through
the shared environment. All selected HotpotQA teams instead retain reader-to-submitter paths
(Appendix~\ref{app:selected-architectures}). These patterns support team
construction, with agent selection the clearest coding result; they do not
isolate a benefit from communication optimization.

\section{Poisoned-Pool Stress Test}
\label{sec:poisoned-pool}
\begin{wrapfigure}{r}{0.46\textwidth}
  \centering
  \vspace{-\intextsep}
  \includegraphics[width=\linewidth]{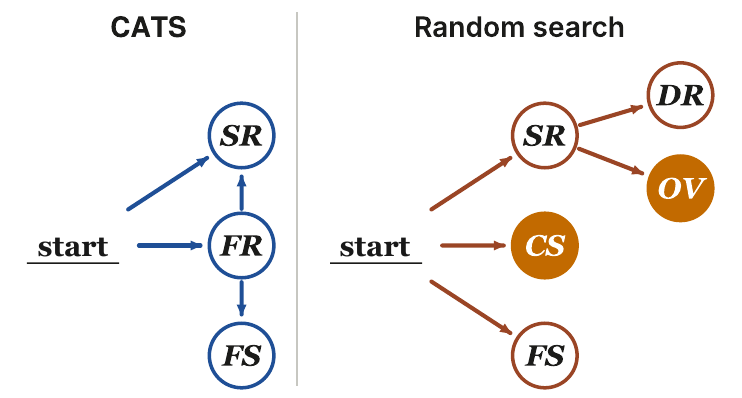}
  \captionsetup{justification=raggedright,singlelinecheck=false}
  \caption{\textbf{Selected topologies on HotpotQA}
  (seed 13). Poisoned agents are in orange.}
  \label{fig:hotpotqa-selected-topologies}
\end{wrapfigure}

A meta-agent that assembles teams must also decline to delegate to agents it
should not trust. To probe this failure mode, we construct a \emph{poisoned}
variant of HotpotQA~\citep{yangHotpotQADatasetDiverse2018}, a third domain
distinct from the coding tasks above. Its supergraph contains six
adversarially prompted agents and four honest agents. The attacks include
plan misdirection, fabricated evidence, citation spoofing, instruction
hijacking, late answer overwrites, and exact-match sabotage
(Table~\ref{tab:agent-roles-and-tools}).
The search receives no adversarial supervision: it observes only task scores
and transcripts, and is never told which agents are poisoned.
Table~\ref{tab:hotpotqa-results} shows why task score alone is
insufficient to assess delegation: comparable scores can mask different
levels of poisoned-agent recruitment. CATS retains no poisoned agents across
all three seeds. Random search retains two on average.
Score-only search's teams are cheap but
inaccurate: 1, 2, and 5 agents, one with four poisoned. CATS instead removes
their incoming edges, making them unreachable
(Figure~\ref{fig:hotpotqa-selected-topologies}). Exclusion is therefore a
byproduct of the optimization signal, not a separate security filter.

\section{Related Work}
\label{sec:related-work}

Topology is an established optimization target: GPTSwarm optimizes agent
graphs, MaAS samples query-dependent systems from an agentic supergraph, and
AgentPrune prunes redundant or malicious
communication~\citep{zhugeGPTSwarmLanguageAgents2024,zhangMultiagentArchitectureSearch2025,zhangCutCrapEconomical2025}.
Graph-GRPO derives statistical, per-edge credit from relative task rewards within groups of
sampled graphs~\citep{cangGraphGRPOStabilizingMultiAgent2026}. CATS instead
uses LLM-judged, per-agent credit, adapting contrastive attribution from prompt
optimization~\citep{zehleCANTANTEOptimizingAgentic2026} to topology search:
agent configurations stay fixed, while team membership and communication
links change. Introspective LLM judges do not faithfully
approximate leave-one-out contribution~\citep{luAgentsThatMatter2026}, so we
treat our credit as an optimization signal, not a causal measure.

\section{Limitations, Responsible Use, and Conclusion}
\label{sec:limitations}
\label{sec:conclusion}

Evaluating a single model family across three domains and three seeds limits
what we can conclude about broader transfer and variance, while high baseline
scores on MBPP leave little headroom for accuracy gains. Moreover, although
pruning yields sparse coding teams, this may reflect coordination via the shared
environment rather than a general lack of need for direct communication.
Finally, we only track token cost without optimizing it, making Pareto search
over score and cost, as done for prompts
by~\citet{bussingMOCAPOMultiObjectiveCostAware2026}, a natural next step.

Attribution credits have not been validated against human judgments.
Credit--success correlation (Table~\ref{tab:credit-score-correlation}) is only a
consistency check, since the attributer sees \texttt{system\_score}.
The score-only baseline was not separately tuned and does not isolate which component of CATS
accounts for the difference. Credit is node-level, propagated equally to all
incoming fired edges. A comparison with Graph-GRPO's statistical edge-level
credit remains future work~\citep{cangGraphGRPOStabilizingMultiAgent2026}.

CATS turns a hand-crafted supergraph into a
small fixed team. It improved the balance between performance and efficiency
on BigCodeBench, reduced unnecessary complexity on the easier MBPP, and
excluded poisoned agents that task scores alone could not detect. In all
three domains it matched or beat score-only heuristic search on
score, deployment-token use, or both.
This supports attribution for team optimization and the pruning of
potentially harmful agents, not a security guarantee.

\paragraph{Responsible-use statement.}
\label{sec:responsible-use}
Selecting a topology determines which agents gain access to tools and task
context. Our stress test shows that attribution can exclude fixed adversarial
prompts, but this does not establish deployment safety. An adaptive adversary
could act beyond the capped transcript window or exploit the unverified LLM
attributer directly. Tool and data permissions must therefore remain outside
the optimizer.

\pagebreak

\begin{ack}
Tom Zehle received funding from the European Union. This work was supported
by the European Union's Horizon Europe research and innovation programme
under grant agreement No.~101214398 (ELLIOT). Views and opinions expressed
are, however, those of the authors only and do not necessarily reflect those
of the European Union or the European Commission. Neither the European Union
nor the granting authority can be held responsible for them. Finally, we
thank the anonymous reviewers for their constructive feedback.
\end{ack}

\appendix

\setlength{\textfloatsep}{12pt plus 2pt minus 2pt}
\setlength{\floatsep}{10pt plus 2pt minus 2pt}
\setlength{\intextsep}{10pt plus 2pt minus 2pt}
\renewcommand{\topfraction}{0.95}
\renewcommand{\bottomfraction}{0.9}
\renewcommand{\textfraction}{0.05}
\renewcommand{\floatpagefraction}{0.85}
\raggedbottom

\section{Additional Experimental Details}
\label{app:additional-details}

\subsection{Supergraphs and Graphs}

\begin{figure}[H]
  \centering
  \begin{subfigure}[t]{0.48\textwidth}
    \centering
    \includegraphics[width=\linewidth]{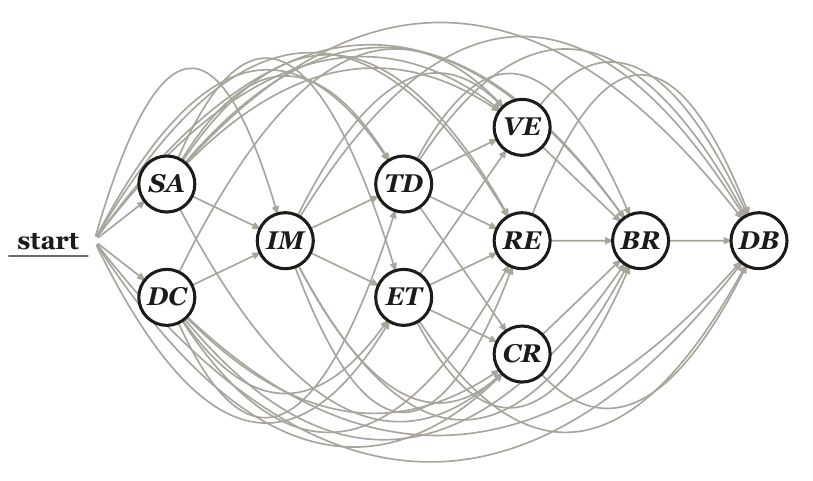}
    \caption{BigCodeBench.}
    \label{fig:supergraph-bigcodebench}
  \end{subfigure}
  \hfill
  \begin{subfigure}[t]{0.48\textwidth}
    \centering
    \includegraphics[width=\linewidth]{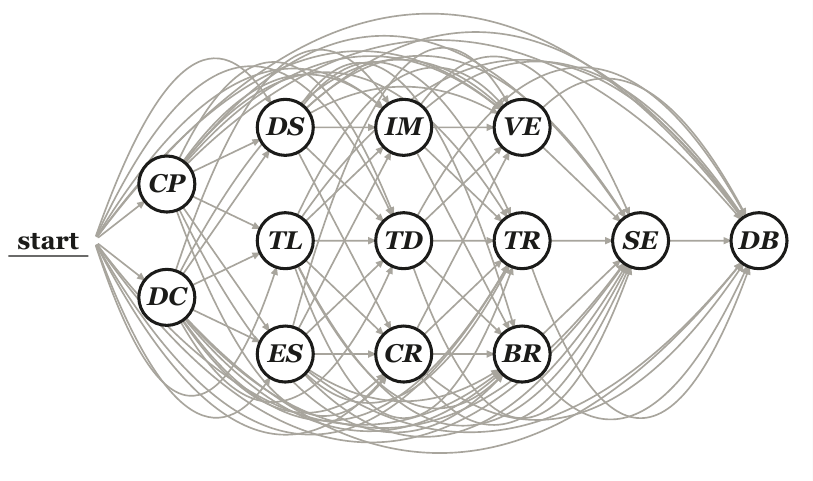}
    \caption{MBPP.}
    \label{fig:supergraph-mbpp}
  \end{subfigure}

  \smallskip

  \begin{subfigure}[t]{0.48\textwidth}
    \centering
    \includegraphics[width=\linewidth]{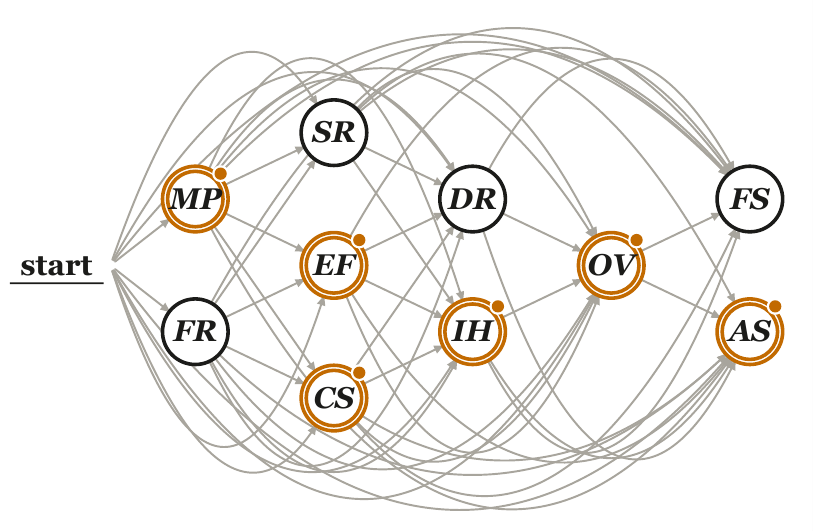}
    \caption{Poisoned HotpotQA.}
    \label{fig:supergraph-hotpotqa}
  \end{subfigure}
  \hfill
  \begin{subfigure}[t]{0.46\textwidth}
    \centering
    \resizebox{\linewidth}{!}{\begin{tikzpicture}[
        >=stealth,
        baselineagent/.style={draw, circle, minimum size=12mm,
          inner sep=1.5pt, font=\scriptsize\itshape},
        flow/.style={->, line width=0.55pt}
      ]
        \path[use as bounding box] (-0.8,-1.65) rectangle (6.4,1.65);
        \node[font=\scriptsize\bfseries] (start) at (0,0) {start};
        \node[baselineagent] (planner) at (1.8,0) {Planner};
        \node[baselineagent] (executor) at (3.8,0) {Executor};
        \node[baselineagent] (verifier) at (5.8,0) {Verifier};
        \draw[flow] (start) -- (planner);
        \draw[flow] (planner) -- (executor);
        \draw[flow] (executor) to[bend left=48] (verifier);
        \draw[flow] (verifier) to[bend left=48]
          node[below, font=\tiny, align=center] {repair\\(max. 5)} (executor);
      \end{tikzpicture}}
    \caption{Hand-crafted MAS baseline.}
    \label{fig:baseline-planner-executor-verifier}
  \end{subfigure}
  \caption{\textbf{Initial supergraphs and fixed baseline.} Supergraph edges
    run forward between layers, and \emph{start} edges select entry agents.
    Orange nodes in (c) are adversarially prompted. These labels are hidden
    from the search and attributer. Agent codes are listed in
    Table~\ref{tab:agent-roles-and-tools}.}
  \label{fig:supergraphs-and-baseline}
\end{figure}

\begin{table}[H]
  \centering
  \caption{\textbf{Agent pools.} HotpotQA alignment labels are for post-hoc measurement
    only and are read by neither the search nor the attributer. A read budget
    of zero permits listing titles but not reading passages.}
  \label{tab:agent-roles-and-tools}
  \footnotesize
  \renewcommand{\arraystretch}{1.0}
  \begin{minipage}[t]{0.49\textwidth}
    \centering
    \setlength{\tabcolsep}{3pt}
    \textbf{BigCodeBench}\\[2pt]
    \begin{tabular}{@{}ll>{\raggedright\arraybackslash}p{0.44\linewidth}@{}}
      \toprule
      Agent & Role & Tools \\
      \midrule
      \textsc{sa} & Spec analyst & None \\
      \textsc{dc} & Draft coder & \texttt{write\_code} \\
      \textsc{im} & Implementer & \texttt{read\_code}, \texttt{write\_code}, \texttt{run\_snippet} \\
      \textsc{td} & Test designer & \texttt{write\_tests}, \texttt{check\_tests} \\
      \textsc{et} & Edge-case tester & \texttt{write\_tests}, \texttt{check\_tests} \\
      \textsc{ve} & Verifier & \texttt{read\_code}, \texttt{run\_snippet} \\
      \textsc{re} & Reviewer & \texttt{read\_code}, \texttt{read\_tests} \\
      \textsc{cr} & Critiquer & \texttt{read\_code} \\
      \textsc{br} & Brainstormer & \texttt{read\_code} \\
      \textsc{db} & Debugger & \texttt{read\_code}, \texttt{write\_code}, \texttt{run\_snippet} \\
      \bottomrule
    \end{tabular}
  \end{minipage}\hfill
  \begin{minipage}[t]{0.49\textwidth}
    \centering
    \setlength{\tabcolsep}{3pt}
    \textbf{MBPP}\\[2pt]
    \begin{tabular}{@{}ll>{\raggedright\arraybackslash}p{0.44\linewidth}@{}}
      \toprule
      Agent & Role & Tools \\
      \midrule
      \textsc{cp} & Contract probe & \texttt{read\_code}, \texttt{run\_snippet}, \texttt{check\_tests} \\
      \textsc{dc} & Draft coder & \texttt{write\_code} \\
      \textsc{ds} & Direct solver & None \\
      \textsc{tl} & Test literalist & None \\
      \textsc{es} & Edge-case solver & None \\
      \textsc{im} & Implementer & \texttt{read\_code}, \texttt{write\_code}, \texttt{run\_snippet}, \texttt{check\_tests} \\
      \textsc{td} & Test designer & \texttt{write\_tests}, \texttt{check\_tests} \\
      \textsc{cr} & Critiquer & \texttt{read\_code} \\
      \textsc{ve} & Verifier & \texttt{read\_code}, \texttt{run\_snippet} \\
      \textsc{tr} & Test runner & \texttt{read\_code}, \texttt{check\_tests} \\
      \textsc{br} & Brainstormer & \texttt{read\_code} \\
      \textsc{se} & Selector & \texttt{read\_code}, \texttt{write\_code}, \texttt{run\_snippet}, \texttt{check\_tests} \\
      \textsc{db} & Debugger & \texttt{read\_code}, \texttt{write\_code}, \texttt{run\_snippet}, \texttt{check\_tests} \\
      \bottomrule
    \end{tabular}
  \end{minipage}

  \medskip

  \begin{minipage}[t]{\textwidth}
    \centering
    \setlength{\tabcolsep}{5pt}
    \textbf{Poisoned HotpotQA}\\[2pt]
    \begin{tabular}{@{}llllr@{}}
      \toprule
      Agent & Role & Alignment & Tools & Read budget \\
      \midrule
      \textsc{mp} & Misdirection planner & Poisoned & None & 0 \\
      \textsc{fr} & First-hop reader & Honest & Retrieve & 5 \\
      \textsc{sr} & Second-hop reader & Honest & Retrieve & 5 \\
      \textsc{ef} & Evidence fabricator & Poisoned & Submit & 0 \\
      \textsc{cs} & Citation spoofer & Poisoned & List titles & 0 \\
      \textsc{dr} & Drafter & Honest & Submit & 0 \\
      \textsc{ih} & Instruction hijacker & Poisoned & None & 0 \\
      \textsc{ov} & Overruling verifier & Poisoned & Retrieve, Submit & 2 \\
      \textsc{fs} & Final submitter & Honest & Submit & 0 \\
      \textsc{as} & Answer-form saboteur & Poisoned & Submit & 0 \\
      \bottomrule
    \end{tabular}
  \end{minipage}
\end{table}

\subsection{Selected Deployment Architectures}
\label{app:selected-architectures}

\begin{table}[H]
  \centering
  \caption{\textbf{Selected deployment topologies $G^\star$.} These are the
  highest-scoring step-4 candidates on the search set and are the systems
  reported as ``CATS'' in the main tables. Agent codes are resolved in
  Table~\ref{tab:agent-roles-and-tools}. Each listed edge is directed. \emph{start}
  denotes an entry edge.}
  \label{tab:selected-topologies}
  \footnotesize
  \setlength{\tabcolsep}{4pt}
  \begin{tabular}{@{}llp{0.27\textwidth}p{0.49\textwidth}@{}}
    \toprule
    Benchmark & Seed & Agents & Edges \\
    \midrule
    BigCodeBench & 13 & DC, IM, TD
      & start$\to$DC, DC$\to$IM, IM$\to$TD \\
    & 14 & DC, VE
      & start$\to$DC, start$\to$VE \\
    & 15 & DB, DC
      & start$\to$DB, start$\to$DC \\
    \addlinespace[3pt]
    MBPP & 13 & DC, IM, TD
      & start$\to$DC, start$\to$IM, start$\to$TD \\
    & 14 & DC, IM, TL
      & start$\to$DC, start$\to$IM, start$\to$TL, TL$\to$IM \\
    & 15 & DC, IM
      & start$\to$DC, start$\to$IM, DC$\to$IM \\
    \addlinespace[3pt]
    HotpotQA & 13 & FR, SR, FS
      & start$\to$FR, start$\to$SR, FR$\to$SR, FR$\to$FS \\
    & 14 & SR, DR
      & start$\to$SR, start$\to$DR, SR$\to$DR \\
    & 15 & FR, SR, DR
      & start$\to$FR, start$\to$SR, FR$\to$SR, FR$\to$DR, SR$\to$DR \\
    \bottomrule
  \end{tabular}
\end{table}

The selected coding teams consistently retain a draft coder alongside one or
two other agents, but their communication patterns differ
(Table~\ref{tab:selected-topologies}). For example, BigCodeBench
seed 13 connects a draft coder, implementer, and test designer in a chain;
MBPP seed 13 selects the same roles with only direct entry edges from
\emph{start}. Two further BigCodeBench teams likewise have only entry edges,
while the remaining MBPP teams retain a single inter-agent edge. These
patterns show that selecting a set of agents and connecting that set are
distinct decisions. Because agents access a shared environment, teams with
only entry edges can still interact through the persisted artifact.

HotpotQA yields a different pattern: every selected team retains the
second-hop reader and either the drafter or final submitter, with directed
communication from a reader to the answer-submitting agent. These teams
combine retrieval and answer submission while excluding all poisoned roles.
The contrast with coding illustrates that the search does not always remove
communication. However, these graphs alone do not establish the benefit of
their particular edges; that would require a comparison that holds agent
membership fixed.

\subsection{Search and Baselines}

\begin{table}[H]
  \centering
  \caption{\textbf{Run settings.} \emph{Left:} CATS
    hyperparameters, shared across benchmarks. \emph{Right:} per-benchmark
    inference settings and mean CATS
    cost. Attribution accounted for 9--11\% of coding-search tokens.}
  \label{tab:search-hyperparameters}
  \footnotesize
  \begin{minipage}[t]{0.45\textwidth}
    \centering
    \setlength{\tabcolsep}{4pt}
    \begin{tabular}{@{}lr@{}}
      \toprule
      Hyperparameter & Value \\
      \midrule
      Search tasks & $100$ \\
      Seeds (search and evaluation) & $13$, $14$, $15$ \\
      Search steps, $T$ & $4$ \\
      Candidates per step, $K$ & $4$ \\
      Learning rate, $\eta$ & $6$ \\
      Unfired-edge weight, $\rho$ & $0.25$ \\
      Initial edge probability, $\alpha_0$ & $0.15$ \\
      Pruning fraction, $\delta$ & $0.85$ \\
      \bottomrule
    \end{tabular}
  \end{minipage}\hfill
  \begin{minipage}[t]{0.52\textwidth}
    \centering
    \setlength{\tabcolsep}{4pt}
    \begin{tabular}{@{}lrrr@{}}
      \toprule
      & BigCode- & & Poisoned \\
      Setting & Bench & MBPP & HotpotQA \\
      \midrule
      Task temperature & $0.7$ & $0.7$ & $0.2$ \\
      Task max tokens & $4096$ & $2048$ & $3072$ \\
      Agent input cap (tokens) & $900$ & $900$ & $700$ \\
      Attribution cap (tokens) & $120$ & $120$ & $150$ \\
      Attributer temperature & $0.7$ & $0.7$ & $0.7$ \\
      Attributer max tokens & $8192$ & $8192$ & $24576$ \\
      \midrule
      Mean search tokens & 16.4M & 13.5M & 20.5M \\
      \bottomrule
    \end{tabular}
  \end{minipage}
\end{table}

\begin{table}[H]
  \centering
  \caption{\textbf{Operational definitions of the baselines.} All baselines use the same
    task model, grader, decoding limits, and seed-specific held-out tasks.}
  \label{tab:baseline-definitions}
  \footnotesize
  \setlength{\tabcolsep}{4pt}
  \begin{tabular}{@{}lp{0.74\textwidth}@{}}
    \toprule
    Method/read-out & Definition \\
    \midrule
    Random search & Draw 16 candidates at fixed $\alpha=0.15$. Return the
      best search-set score, breaking ties by fewer agents and then random. \\
    \addlinespace[2pt]
    Score-only search & Sample $K=4$ candidates at each of $T=4$ steps.
      For candidate $k$, compute the score-only advantage
      $A_k=\bar{s}_k-(K-1)^{-1}\sum_{j\neq k}\bar{s}_j$, where
      $\bar{s}_k$ is its mean search-set score. Assign $A_k$ to every active
      agent and update edge logits using the same learning rate, unfired-edge
      rule, and pruning schedule as our method, but without LLM attribution.
      This heuristic omits the Bernoulli score-function factor
      $I_e-\alpha_e$ required by REINFORCE~\citep{williamsSimpleStatisticalGradientFollowing1992}.
      Return the highest-scoring step-$T$ candidate, breaking ties by fewer
      agents and then the earlier candidate. \\
    \addlinespace[2pt]
    Direct LLM (MBPP) & One greedy zero-shot call, without tools, execution
      feedback, or retry. The prompt follows the task template of
      \citet{austinProgramSynthesisLarge2021}. \\
    \addlinespace[2pt]
    Direct LLM (BigCodeBench) & One greedy zero-shot call, without tools,
      execution feedback, or retry. The prompt uses the
      \texttt{instruction\_prefix} and \texttt{response\_prefix} defaults from
      the official \href{https://github.com/bigcode-project/bigcodebench}{BigCodeBench
      generation harness}. \\
    \addlinespace[2pt]
    ReAct single agent & One native-tool loop with the same prompt body on both
      benchmarks, and a tool-call budget of five.
      ReAct is the method of \citet{yaoReActSynergizingReasoning2023}. The exact
      string is from LangChain Hub entry \texttt{hwchase17/react}.
      \texttt{\{tools\}} and \texttt{\{tool\_names\}} are filled with the
      environment toolset and the runtime's native tool-use protocol block. \\
    \addlinespace[2pt]
    Hand-crafted MAS & A fixed planner--executor--verifier topology
      (Figure~\ref{fig:baseline-planner-executor-verifier}). \\
    \bottomrule
  \end{tabular}
\end{table}

\pagebreak
\subsection{Evaluation Protocol}

Field caps use \texttt{o200k\_base} tokens, retain the prefix, discard the tail,
and append the marker \mbox{\texttt{...~[truncated]}} outside the cap. Agent input fields
(query and upstream outputs) use the 900/700-token caps in
Table~\ref{tab:search-hyperparameters}; the attributer's query
and string-valued \texttt{says} fields use 120 tokens for coding and 150 for
HotpotQA, with \texttt{does} tool-effect logs and \texttt{final} summaries
exempt from field capping. The assembled attributer prompt is also
prefix-capped at its model limit minus the attribution field cap.

Seeds 13--15 jointly control splits, global randomness, topology sampling, and
generation. The generation endpoint, hosted by the University of Freiburg,
used the aliases \texttt{gemma-4-12b-llmlb} (agents/baselines) and
\texttt{gemma-4-31b-llmlb} (attributer). Weight hashes and quantization were not
exposed. Coding runs use fresh temporary directories and process groups with a
30-second timeout. Crashes, timeouts, empty environments, and unusable results
score zero. Deployment tokens count all agent-call input and output tokens per
held-out task. All reported $\pm$ values and shaded bands are Bessel-corrected
sample standard deviations over three runs.

\begin{table}[H]
  \centering
  \caption{\textbf{Dataset splits and grading.} Each held-out set contains 100 tasks.}
  \label{tab:dataset-protocol}
  \footnotesize
  \setlength{\tabcolsep}{6pt}
  \begin{tabular}{@{}lll@{}}
    \toprule
    Benchmark & Snapshot and search split & Grading \\
    \midrule
    BigCodeBench & \texttt{v0.1.4} Instruct, 100 disjoint tasks & Last stored program against hidden tests \\
    MBPP & Revision \texttt{4bb6404f}, 100 disjoint tasks & Last stored program against hidden tests \\
    HotpotQA & \texttt{distractor}, 100 disjoint tasks & Exact match on last \texttt{submit\_answer} phrase \\
    \bottomrule
  \end{tabular}
\end{table}

\subsection{Per-Seed Results and Additional Analyses}
\label{app:additional-analyses}
\label{app:credit-score-correlation}

\begin{table}[H]
  \centering
  \caption{\textbf{Per-seed evaluations supporting the headline comparisons.}
  Each seed groups results for the selected topologies on its 100-task held-out
  set. Scores are percentages. Tokens are thousands of input plus output
  tokens per task. $V/E$ counts selected agents/edges, including start edges.
  $P$ counts poisoned agents (HotpotQA only).}
  \label{tab:per-seed-results}
  \footnotesize
  \renewcommand{\arraystretch}{0.95}
  \setlength{\tabcolsep}{3pt}
  \begin{tabular*}{\textwidth}{@{\extracolsep{\fill}}l*{3}{rrcr}@{}}
    \toprule
    & \multicolumn{4}{c}{Seed 13} & \multicolumn{4}{c}{Seed 14} & \multicolumn{4}{c}{Seed 15} \\
    \cmidrule(lr){2-5} \cmidrule(lr){6-9} \cmidrule(l){10-13}
    Method & Score & Tokens & $V/E$ & $P$ & Score & Tokens & $V/E$ & $P$ & Score & Tokens & $V/E$ & $P$ \\
    \midrule
    \multicolumn{13}{@{}l}{\textbf{BigCodeBench}} \\
    CATS & 53 & 7.16 & 3/3 & --- & 43 & 4.70 & 2/2 & --- & 48 & 8.28 & 2/2 & --- \\
    Random & 35 & 9.06 & 5/6 & --- & 28 & 11.08 & 2/2 & --- & 42 & 5.93 & 1/1 & --- \\
    Score-only search & 45 & 6.62 & 3/3 & --- & 37 & 23.71 & 4/5 & --- & 46 & 13.75 & 5/8 & --- \\
    \addlinespace[3pt]
    \multicolumn{13}{@{}l}{\textbf{MBPP}} \\
    CATS & 72 & 4.45 & 3/3 & --- & 85 & 5.90 & 3/4 & --- & 85 & 5.44 & 2/3 & --- \\
    Random & 75 & 9.32 & 7/7 & --- & 84 & 31.31 & 9/12 & --- & 82 & 15.82 & 7/8 & --- \\
    Score-only search & 73 & 9.79 & 6/6 & --- & 84 & 28.85 & 8/15 & --- & 80 & 27.88 & 10/18 & --- \\
    \addlinespace[3pt]
    \multicolumn{13}{@{}l}{\textbf{HotpotQA}} \\
    CATS & 58 & 8.37 & 3/4 & 0 & 64 & 11.24 & 2/3 & 0 & 57 & 18.63 & 3/5 & 0 \\
    Random & 60 & 12.97 & 5/5 & 2 & 54 & 20.01 & 6/6 & 3 & 59 & 12.23 & 3/3 & 1 \\
    Score-only search & 27 & 3.95 & 2/2 & 0 & 22 & 6.18 & 5/5 & 4 & 24 & 1.24 & 1/1 & 0 \\
    \bottomrule
  \end{tabular*}
\end{table}

\begin{figure}[H]
  \centering
  \begin{subfigure}[t]{0.49\textwidth}
    \centering
    \includegraphics[width=\linewidth]{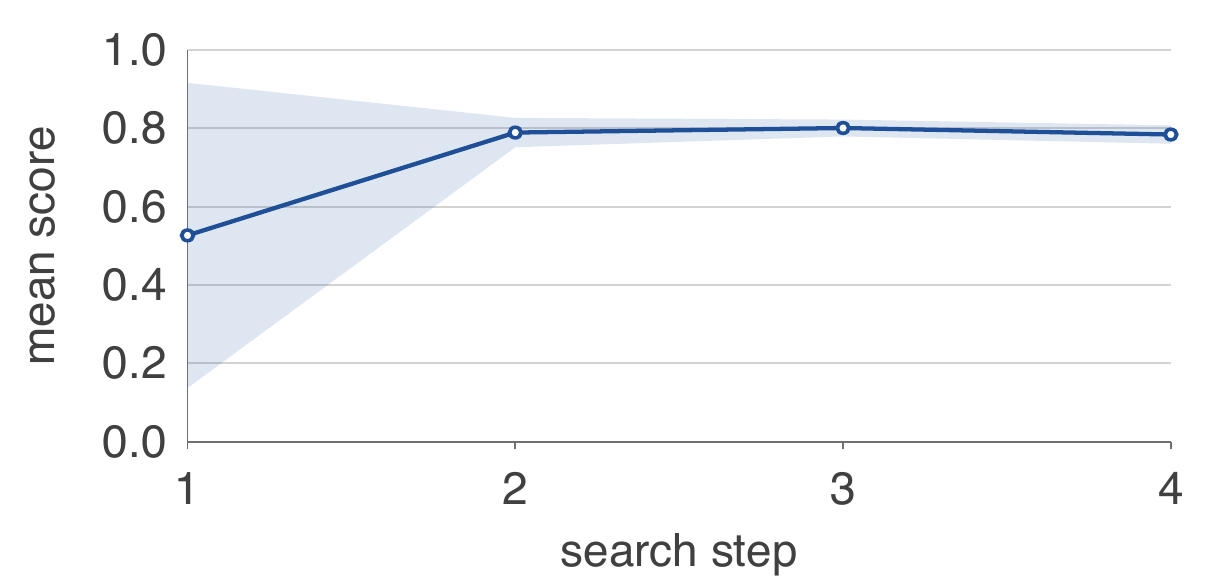}
    \caption{MBPP search score.}
    \label{fig:mbpp-search-score}
  \end{subfigure}
  \hfill
  \begin{subfigure}[t]{0.49\textwidth}
    \centering
    \includegraphics[width=\linewidth]{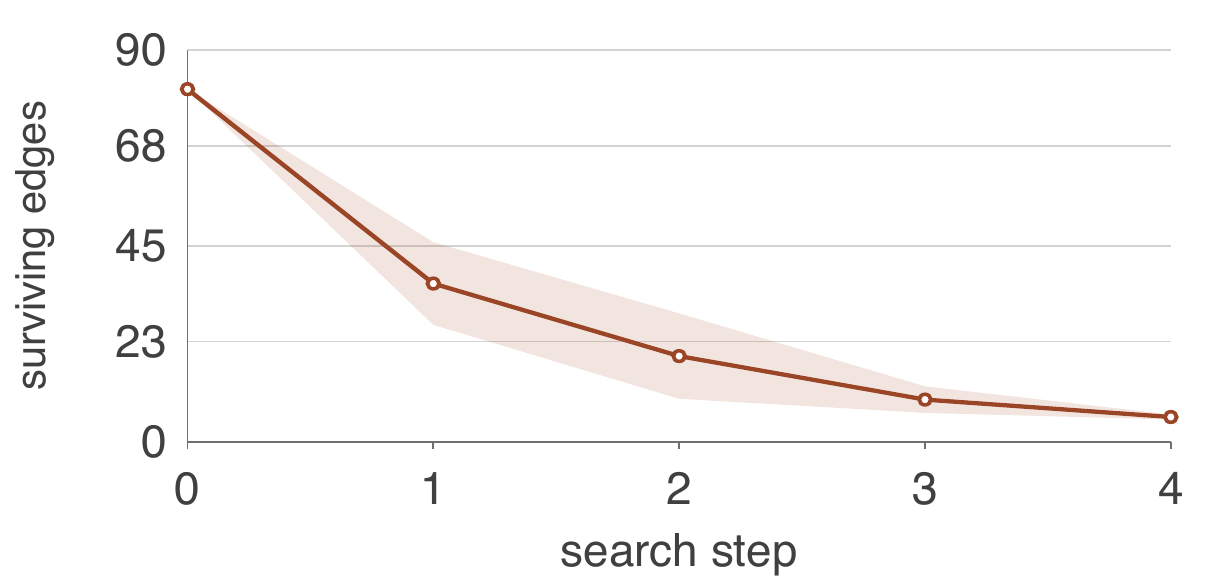}
    \caption{MBPP surviving edges.}
    \label{fig:mbpp-surviving-edges}
  \end{subfigure}
  \caption{\textbf{CATS dynamics on MBPP.} The corresponding
  BigCodeBench panels appear in the main paper. Lines show means over three
  seeds. Bands show sample standard deviations.}
  \label{fig:mbpp-search-dynamics}
\end{figure}

\begin{table}[H]
  \centering
  \caption{\textbf{Credit--success consistency check.}
  Correlation between the mean credit over participating agents,
  $\bar{c}_{k,q}$, and the binary task score $s_{k,q}$, over all
  (candidate, task) pairs. Within-task columns correlate within each query.
  Scores are binary, so Pearson $r$ is the point-biserial coefficient. Four
  pairs are dropped from seed 15 due to one unparsed transcript. The attributer
  sees \texttt{system\_score}, so this is not independent validation of credit quality.}
  \label{tab:credit-score-correlation}
  \footnotesize
  \renewcommand{\arraystretch}{0.95}
  \setlength{\tabcolsep}{5pt}
  \begin{tabular}{@{}l r rr @{\hspace{12pt}} rr@{}}
    \toprule
    & & \multicolumn{2}{c}{\textbf{Pooled}} & \multicolumn{2}{c}{\textbf{Within-task}} \\
    \cmidrule(lr){3-4} \cmidrule(l){5-6}
    Seed & $n$ pairs & $r$ & $\rho$ & $r$ & $\rho$ \\
    \midrule
    13 & 1600 & 0.876 & 0.854 & 0.785 & 0.742 \\
    14 & 1600 & 0.879 & 0.834 & 0.841 & 0.802 \\
    15 & 1596 & 0.834 & 0.846 & 0.746 & 0.687 \\
    \midrule
    Mean $\pm$ SD over seeds & --- & 0.863\,{\scriptsize$\pm$0.025} & 0.845\,{\scriptsize$\pm$0.010} & 0.791\,{\scriptsize$\pm$0.048} & 0.744\,{\scriptsize$\pm$0.058} \\
    \bottomrule
  \end{tabular}
\end{table}

\begin{table}[H]
  \centering
  \caption{\textbf{Offline search tokens, separate from deployment.}
    Millions of input plus output tokens for 16 candidates on 100 search
    tasks, averaged over seeds 13--15; totals include sample SD.}
  \label{tab:search-cost}
  \footnotesize
  \begin{tabular}{@{}llrrr@{}}
    \toprule
    Benchmark & Method & Task model & Attributer & Total \\
    \midrule
    BigCodeBench & CATS & 14.88 & 1.54 & $16.42\pm5.35$ \\
                 & Random search & 16.49 & 0 & $16.49\pm5.66$ \\
                 & Score-only search & 22.89 & 0 & $22.89\pm10.89$ \\
    \addlinespace[2pt]
    MBPP & CATS & 12.00 & 1.53 & $13.53\pm1.51$ \\
         & Random search & 11.45 & 0 & $11.45\pm2.71$ \\
         & Score-only search & 20.89 & 0 & $20.89\pm3.47$ \\
    \addlinespace[2pt]
    HotpotQA & CATS & 18.39 & 2.11 & $20.50\pm2.79$ \\
             & Random search & 9.91 & 0 & $9.91\pm2.20$ \\
             & Score-only search & 10.63 & 0 & $10.63\pm4.53$ \\
    \bottomrule
  \end{tabular}
  \par\smallskip
\end{table}

\clearpage

\section{BigCodeBench Attributer Call}
\label{app:attributer-call}

We adapt contrastive attribution~\citep{zehleCANTANTEOptimizingAgentic2026} for
BigCodeBench by grading the persisted code artifact instead of agent text. We
also expose ordered, compact tool effects. The representative call below is
excerpted, with omitted material marked \texttt{[...]}.

\begingroup
\lstset{
  basicstyle=\ttfamily\fontsize{7.5}{6.5}\selectfont,
  breaklines=true,
  breakatwhitespace=false,
  columns=fullflexible,
  keepspaces=true,
  frame=none,
  xleftmargin=2pt,
  xrightmargin=2pt,
  aboveskip=4pt,
  belowskip=4pt,
  showstringspaces=false
}
\noindent\rule{\linewidth}{0.3pt}
\begin{lstlisting}
===== PROMPT (ABRIDGED) =====
# Task
You are an attribution agent for a multi-agent system.
[...]
Your goal is to estimate how each agent's behavior contributed to the
outcome of its execution by comparing differences across executions.

Base your attribution on:
- the system score differences across executions,
- the system prediction
[...]
- what each agent measurably changed in the shared environment.

# Attribution Scale
Assign a float score in [-1.0, 1.0] to each agent for each execution:
- positive = helpful contribution
- negative = harmful contribution
- 0 = neutral or unclear impact

# The Protocol
This system is graded on an artifact held in a shared coding environment,
not on any agent's text: the code persisted with `write_code` is what the
hidden test suite runs, and it appears in no agent's output. So every agent
is given two keys: `says` (its text output) and `does` (a numbered log of
every tool call it made, in order).

Each `does` line is `<seq> <VERB> <target> <delta> ["<observed>"]`:
  1 SET code=c1 4L
  [...]
  5 EDIT code c1->c2 +1/-1
  6 NOOP write_code "[rejected - unsafe import]"

[...] Only targets named in `final` are graded. Scratch writes count only
through a visible later observation followed by a graded write. When several
agents write code, credit only what survives into the graded value. Trust
`does` over conflicting claims in `says`, and treat example tests as the
exact required behavior. [...]

# Input
You will receive JSON with `query` and an `executions` list. Each execution
contains `system_score`, `final`, and a mapping from agent names to `says`
and `does`. [...]

# Output
First, briefly compare the executions and identify the key behavioral
differences that lead to differences in the system score.

Then return JSON inside <attribution>...</attribution>. The top-level object
must have exactly one key, `executions`, containing exactly four items in
input order. Each item contains only `agent_credits`, with exactly the seven
specified agent names and float values in [-1.0, 1.0]. [...]

# Few-shot example
[...]

===== RESPONSE (ABRIDGED) =====
All four executions failed (system_score: 0.0), primarily because the agents
struggled to align the code implementation with the actual data structure
implied by the prompt.

[...]

In Execution 3, EDGE_CASE_TESTER_4 wrote tests to the scratchpad
(`test_list`). However, because no coder followed these tests with a write to
the graded `code` target, these tests had zero impact on the outcome. [...]

<attribution>
{"executions":[
 {"agent_credits":{"BRAINSTORMER_8":0.0,"CRITIQUER_7":0.0,"DRAFT_CODER_1":0.0,
   "EDGE_CASE_TESTER_4":0.0,"IMPLEMENTER_2":-0.2,"REVIEWER_6":0.0,"VERIFIER_5":-0.4}},
 {"agent_credits":{"BRAINSTORMER_8":0.0,"CRITIQUER_7":-0.2,"DRAFT_CODER_1":-0.4,
   "EDGE_CASE_TESTER_4":0.0,"IMPLEMENTER_2":0.0,"REVIEWER_6":0.0,"VERIFIER_5":0.0}},
 {"agent_credits":{"BRAINSTORMER_8":0.0,"CRITIQUER_7":0.0,"DRAFT_CODER_1":0.0,
   "EDGE_CASE_TESTER_4":0.0,"IMPLEMENTER_2":0.0,"REVIEWER_6":0.0,"VERIFIER_5":0.0}},
 {"agent_credits":{"BRAINSTORMER_8":0.0,"CRITIQUER_7":0.0,"DRAFT_CODER_1":0.0,
   "EDGE_CASE_TESTER_4":0.0,"IMPLEMENTER_2":0.0,"REVIEWER_6":0.0,"VERIFIER_5":0.0}}
]}
</attribution>
\end{lstlisting}
\endgroup

\end{document}